\documentclass[unnumsec,webpdf,contemporary,large,numbered]{oup-authoring-template} 

\makeatletter
\global\@numbibtrue
\makeatother

\graphicspath{{Fig/}}

\theoremstyle{thmstyleone}%
\theoremstyle{thmstyletwo}%
\theoremstyle{thmstylethree}%

\usepackage{booktabs}

\begin{document}

\journaltitle{Journal}
\DOI{DOI added during production}
\copyrightyear{2026}
\pubyear{2026}
\vol{XX}
\issue{x}
\access{Published: Date added during production}
\appnotes{Problem Solving Protocol}

\firstpage{1}


\title[Short Article Title]{CrossADR: enhancing adverse drug reactions prediction for combination pharmacotherapy with cross-layer feature integration and cross-level associative learning}

\author[1,$\ast$]{Y. Cheung}

\address[1]{\orgdiv{School of Mathematical Sciences and Shanghai Key Laboratory of PMMP}, \orgname{East China Normal University}, \orgaddress{\postcode{200241}, \state{Shanghai}, \country{China}}}

\corresp[$\ast$]{Corresponding author. \href{email:52275500001@stu.ecnu.edu.cn}{52275500001@stu.ecnu.edu.cn}}

\received{Date}{0}{Year}
\revised{Date}{0}{Year}
\accepted{Date}{0}{Year}


\abstract{Combination pharmacotherapy offers substantial therapeutic advantages but also poses substantial risks of adverse drug reactions (ADRs). The accurate prediction of ADRs with interpretable computational methods is crucial for clinical safety management, drug development, and precision medicine. However, managing ADRs remains a challenge due to the vast search space of drug combinations and the complexity of physiological responses. Current graph-based architectures often struggle to effectively integrate multi-scale biological information and frequently rely on fixed association matrices, which limits their ability to capture dynamic organ-level dependencies and generalize across diverse datasets. Here we propose CrossADR, a hierarchical framework for organ-level ADR prediction through cross-layer feature integration and cross-level associative learning. It incorporates a gated-residual-flow graph neural network to fuse multi-scale molecular features and utilizes a learnable ADR embedding space to dynamically capture latent biological correlations across 15 organ systems. Systematic evaluation on the newly constructed CrossADR-Dataset—covering 1,376 drugs and 946,000 unique combinations—demonstrates that CrossADR consistently achieves state-of-the-art performance across 80 distinct experimental scenarios and provides high-resolution insights into drug-related protein protein interactions and pathways. Overall, CrossADR represents a robust tool for cross-scale biomedical information integration, cross-layer feature integration as well as cross-level associative learning, and can be effectively utilized to prevent ADRs in clinical decision-making.}


\maketitle

\section{Introduction}

Combination pharmacotherapy has emerged as a cornerstone of modern medicine due to its ability to provide complementary efficacy, alleviate toxicity, and delay the onset of drug resistance \cite{weinstein2018modeling, khunsriraksakul2022integrating, iwasaki2020potential}. However, as compounds with inherent bioactivity, drugs often carry the risk of adverse drug reactions (ADRs), which are frequently exacerbated when multiple agents are administered together \cite{bg1, bg2, bg3}. The accurate prediction of ADRs with interpretable computational methods is crucial for clinical safety management, drug development, and precision medicine \cite{wan2025deepadr, gao2025precision, o2023genetic}. Recently, OrganADR pioneered the ADRs study within drug-drug interaction research by emphasizing an organ perspective \cite{OrganADR}. This shifted the focus from broad ADRs to localized physiological impacts, providing a crucial foundation for precise and manageable clinical safety assessments \cite{chen2025evaluation, wan2024multi}.

Despite these conceptual advances, managing ADRs remains a monumental challenge due to the vast search space of drug combinations. With over 2,000 FDA-approved drugs, the potential pairings exceed one million, making it impossible to rely solely on clinical records or the experience of individual physicians across different medical departments \cite{tian2025ddinter, harrison2012predicting}. While computational methods have continuously improved, reports from OrganADR indicate that ADR prediction accuracy at the organ level still hovers between 70\% and 80\%. This gap remains a substantial barrier to meeting the rigorous safety demands of clinical practice and drug management.

Current mainstream architectures typically rely on Knowledge Graphs and Graph Neural Networks (GNN-KG) \cite{li2026mhafr, zhang2026transformer}, yet they often overlook the natural physiological connections between organs. Although OrganADR addressed this by using an ADR association matrix, this approach relies heavily on large, unbiased training data. In real-world scenarios, such models risk learning data biases rather than true, interpretable biological information \cite{li2025llm, ren2025predicting}. Furthermore, while some studies have identified molecular-level and organ-level information within biomedical networks, they usually treat these layers as isolated modules linked only by simple attention mechanisms \cite{su2022attention, zhao2023identifying}. Inspired by residual connections in computer science and biochemical cascades in life sciences, we argue that vital information exists across different scales and stages that previous models have failed to integrate fully. This lack of deep feature fusion serves as the primary computational motivation for our current study.

\begin{figure*}[!t]
\centering
\includegraphics[width=0.80\textwidth]{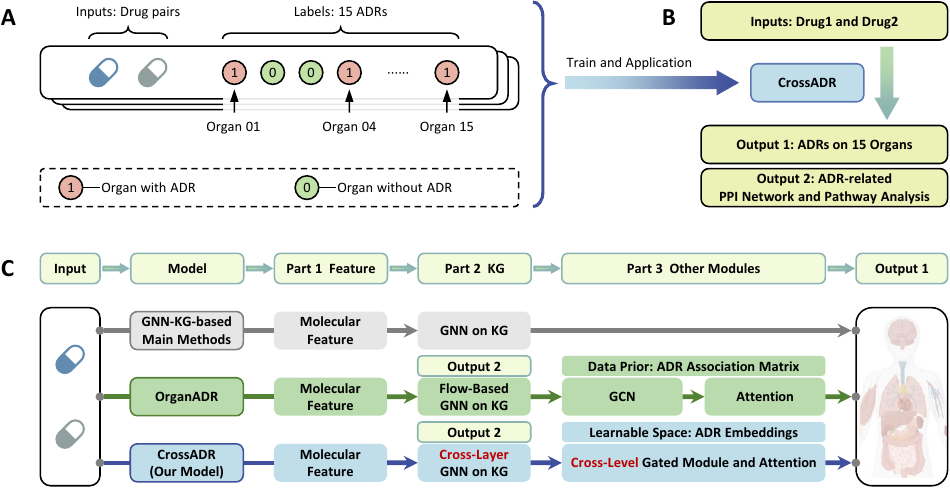}
\caption{Overview of the CrossADR framework for organ-level adverse drug reaction (ADR) prediction. (A) Input: drug pairs, as well as the 15 ADR labels at organ level. (B) The training and application workflow of CrossADR for predicting ADRs and performing downstream PPI network and pathway analysis. (C) Architectural comparison between baseline methods (GNN-KG-based, OrganADR) and the proposed CrossADR model, highlighting two proposed important module: (1) the cross-layer feature integration module, (2) the learnable ADR embeddings as well as cross-level gated module and attention mechanisms.}\label{fig1}
\end{figure*}

To address these gaps, CrossADR, a framework designed for organ-level ADR prediction through cross-layer feature integration and cross-level associative learning, is presented (Fig.\ref{fig1}). The architecture utilizes four types of drug features processed through learnable self-attention modules. Specifically, it employs a cross-layer GNN on Knowledge Graphs that incorporates drug-dependent attention scores on edge types, a gated-residual-flow mechanism, and bi-directional cross-attention. Unlike previous methods, CrossADR abandons fixed association matrices in favor of a learnable space for ADR embeddings. This innovation allows the model to capture organ-level information dynamically, reducing sensitivity to data bias and improving generalization across different datasets. By focusing on the cross-layer fusion of molecular features within the biomedical network and cross-level integration between molecular and organ responses, the model achieves a more robust and associative learning process.

CrossADR is evaluated using the newly constructed CrossADR-Dataset, which covers 1,376 drugs and a vast combination space of 946,000 unique pairs, representing a significant expansion over previous benchmarks. Our evaluation spanned ten independent datasets across 80 distinct scenarios and 15 organ systems. Results show that CrossADR consistently achieves superior performance over both state-of-the-art deep learning models and widely used machine learning methods. Ablation studies further confirm the effectiveness of our architectural innovations, particularly the gated modules and multi-scale fusion. Finally, case studies demonstrate that CrossADR can identify drug-related and ADR-related protein interactions and pathways \cite{jjingo2026pathways}. Given the massive scale of potential drug combinations and the large patient populations affected by multi-morbidity, the performance gains and biological insights provided by CrossADR offer significant value for clinical drug management and patient safety.

\section{Methods}\label{sec2}

\subsection{Problem formulation of ADR prediction}

Following ref.\cite{OrganADR}, the objective of this study is to predict the potential adverse drug reactions (ADRs) that arise when two specific drugs, denoted as $p$ and $q$, are administered together. As illustrated in Fig.\ref{fig1}a-b, the set of ADRs for this drug pair is defined as follows:
\begin{equation}
    \mathcal{A}_{p,q} = \{a^{i}_{p,q}\}, \quad i = 1,2, \ldots, 15,
\end{equation}
In this formulation, the disordered collection of side effects observed under the combined pharmacotherapy of $p$ and $q$ is represented by $\mathcal{A}_{p,q}$, where the index $i$ corresponds to a specific organ system.

For each individual organ, the occurrence of an ADR is captured by a binary value. The specific status of an ADR is determined by:
\begin{equation}
    a^{i}_{p,q} = \begin{cases}
        1, & \text{if an ADR is caused in organ \(i\) by \(p + q\)} \\
        0, & \text{if no ADR is caused in organ \(i\) by \(p + q\)}
    \end{cases}.
\end{equation}
where value of 1 is assigned if an ADR is triggered. Conversely, a value of 0 is recorded if the organ remains unaffected by \(p + q\).

\begin{figure*}[!t]
\centering
\includegraphics[width=0.80\textwidth]{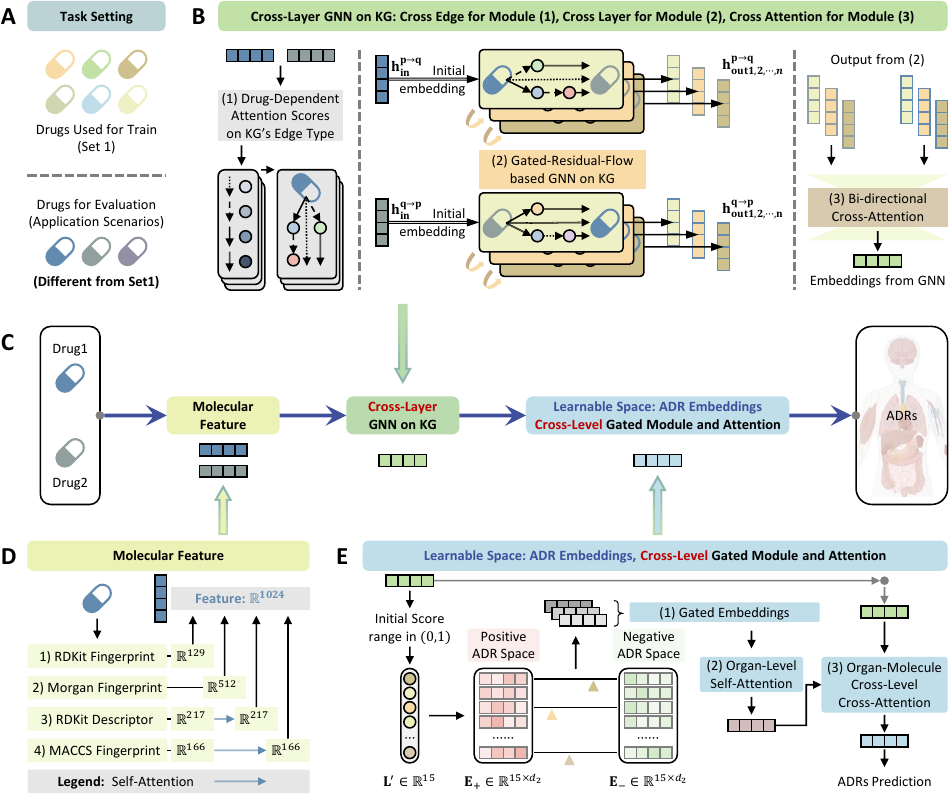}
\caption{Detailed architecture of CrossADR. (A) Task setting for "emerging drug
-emerging drug`` ADR prediction task. (B) The cross-layer GNN on KG module. (C) The workflow of CrossADR. (D) Molecular feature. (E) The cross-level gated module and attention module with learnable ADR embedding space.}\label{fig2}
\end{figure*}

CrossADR focus on "Predict ADRs between two emerging drugs``, with the same task formulation in \cite{OrganADR, EmerGNN}, as shown in Fig.\ref{fig2}a.

\subsection{Model: architecture of CrossADR}

In summary, the architecture of CrossADR establishes a hierarchical information processing pipeline that transitions from microscopic molecular structures to macroscopic organ-level responses. By integrating diverse drug features with prior knowledge from the PrimeKG, the model employs a gated-residual-flow GNN and bi-directional cross-attention to capture the multi-scale evolution of drug-drug interactions. Furthermore, the introduction of a learnable ADR embedding space and cross-level attention mechanisms enables the model to adaptively fuse molecular-level network signals with latent biological correlations between organ systems. This synergistic design ensures that CrossADR can model the complex, non-linear dependencies inherent in multi-organ ADR predictions, providing a robust and interpretable framework for drug safety assessment.

\subsubsection{Molecular feature}

The innovation of CrossADR lies in its architecture. Therefore, for drug representation, CrossADR applies the same features as OrganADR\cite{OrganADR}. The extraction process begins with the generation of initial feature vectors for each drug, denoted as $\mathbf{f}_p \in \mathbb{R}^{1024}$ and $\mathbf{f}_q \in \mathbb{R}^{1024}$, as shown in Fig.\ref{fig2}d. In this phase, one molecular descriptor and three distinct types of molecular fingerprints are extracted. Since the components of the RDKit descriptors and MACCS fingerprints represent specific physical or chemical properties—such as molecular weight or the count of carbon atoms—two independent, trainable self-attention modules are utilized to process them. 

For each feature type $\mathbf{x}_k$ that requires attention, a weight vector $\mathbf{w}_k$ is calculated to capture the importance of different dimensions:
\begin{equation}
    \mathbf{w}_k = \text{Softmax}(\mathbf{W}_k \mathbf{x}_k), \quad \mathbf{e}_k = \mathbf{x}_k \odot \mathbf{w}_k,
\end{equation}
where $\mathbf{W}_k$ denotes the trainable weight matrix of the linear layer, and $\odot$ represents the Hadamard product used for element-wise weighting. 

Specifically, four kinds of drug features applied for CrossADR are:
\begin{itemize}
    \item Continuous molecular descriptors are calculated via the ``CalcMolDescriptors" function in RDKit. These descriptors are used to quantify essential physicochemical properties, such as the molecular weight, and the polar surface area (PSA).
    
    \item Path-based fingerprints are generated using the ``RDKit Fingerprint" function. This is included because the structure and substructure of drugs are identified as critical factors for predicting drug-drug interactions in several studies\cite{nyamabo2021ssi, yu2022stnn, zhu2022molecular}.
    
    \item MACCS keys are employed to provide a clear characterization of key molecular substructures. Through this method, features such as ring structures, element types, and the presence of rare atoms are explicitly encoded into one-hot vectors.
    
    \item Morgan fingerprints (radius is set to 2) are utilized to represent the local atomic environment. The molecular structure is encoded by considering the specific neighborhood around each atom, ensuring that subtle structural differences are captured.
\end{itemize}

\subsubsection{Knowledge graph}
\begin{figure*}[!t]
\centering
\includegraphics[width=0.80\textwidth]{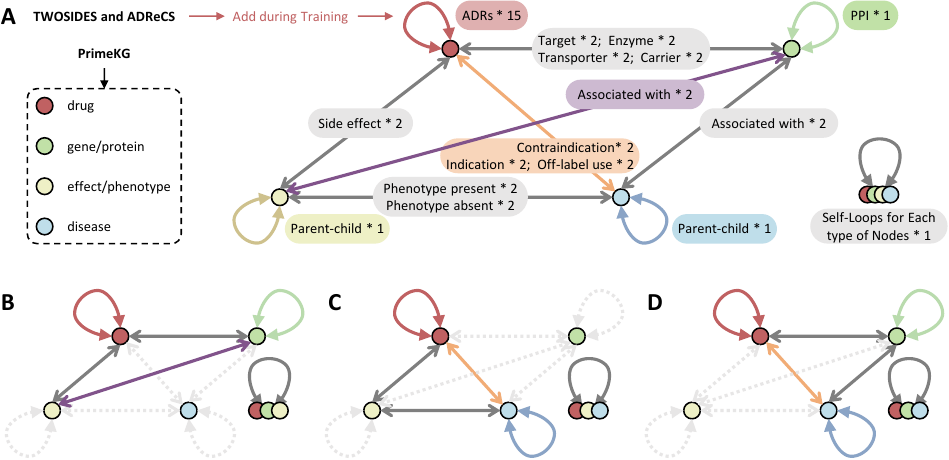}
\caption{Knowledge Graph (KG) construction and ablated KGs. (A) Integration of TWOSIDES and ADReCS, as well as the topology of the basic KG. (B,C,D) Three ablated KGs for comparison. B, C and D for ablated KG 1, 2 and 3. Based on the basic KG, different type of information is removed accordingly.}\label{fig3}
\end{figure*}

Knowledge graph provides the foundational data structure for models' graph convolutional modules and also provides prior biomedical network knowledge for the ADR prediction task. As illustrated in Fig.\ref{fig3}, four node types from PrimeKG are retained, along with their corresponding edges. To ensure the integrity of the evaluation, "synergistic interaction" edges between drugs are removed from PrimeKG to prevent knowledge leakage. Furthermore, bi-directional connections in PrimeKG are filtered into uni-directional interactions that are biologically justified. Detailed edge types and statistics for the knowledge graphs are provided in Table.\ref{tab:kg}.

\begin{figure*}[!t]
\centering
\includegraphics[width=0.80\textwidth]{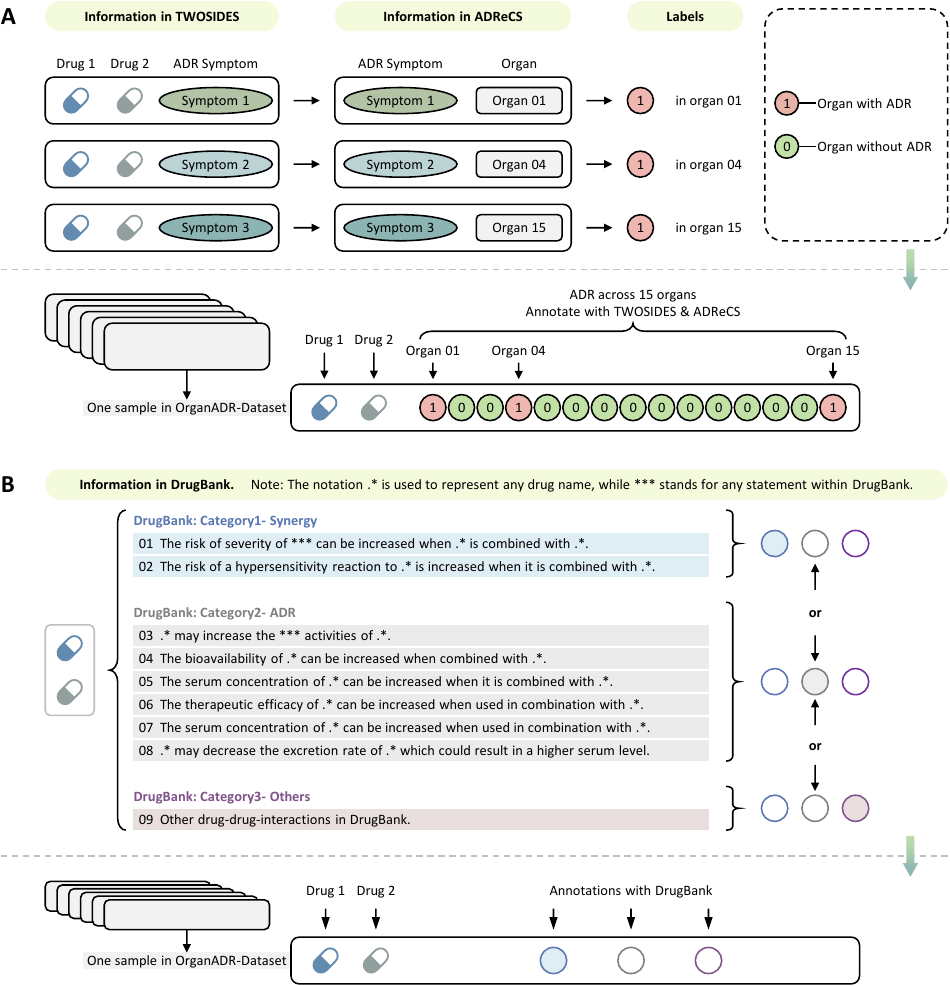}
\caption{Data annotation and sample generation for the CrossADR-Dataset. (A) Mapping drug combinations from TWOSIDES and ADRECS to 15 specific organs to create multi organ ADR labels. (B) Integration of synergistic and ADR-related information from DrugBank to define positive and negative samples.}\label{fig4}
\end{figure*}

\subsubsection{GNN module: Cross-layer GNN on KG}
Following ref.\cite{EmerGNN, OrganADR}, a flow-based GNN module, which represents a state-of-the-art architecture in DDI prediction, is employed to generate embedding vectors for each drug pair. However, CrossADR innovatively captures features more comprehensively through (1) drug-dependent attention scores on KG edge types, (2) a gated-residual-flow-based GNN on the KG, and (3) bi-directional cross-attention, as illustrated in Fig.\ref{fig2}b. Let $\mathbf{f}_p$ and $\mathbf{f}_q$ to represent the molecular features extracted for drugs $p$ and $q$, respectively, the output of the cross-layer GNN module can be summarized as follows:
\begin{equation}
    \mathbf{h}_{out1}^{p, q} = f_\text{GNN}(\mathcal{G}, \mathbf{f}_p, \mathbf{f}_q).
\end{equation}

The cross-layer GNN propagates information across $L$ layers. For each layer $l \in \{1, \dots, L\}$, the module first computes drug-dependent relation attention weights $\alpha_{r}^{(l)}$ based on the drug pair context:
\begin{equation}
    \alpha_{r}^{(l)} = \sigma\left( \mathbf{W}_{attn}^{(l)} \cdot \text{ReLU}\left( \mathbf{W}_{rel}^{(l)} [\mathbf{f}_p ; \mathbf{f}_q] \right) \right),
\end{equation}
where $\mathbf{W}_{rel}^{(l)}$ and $\mathbf{W}_{attn}^{(l)}$ are learnable weight matrices, which scale the relation embeddings $\mathbf{r}^{(l)}$ to form the relation-aware message:
\begin{equation}
    \hat{\mathbf{r}}^{(l)} = \alpha_{r}^{(l)} \odot \mathbf{r}^{(l)}.
\end{equation}

In the $l$-th layer of the GNN module, the representation of an entity $e$ in the flow starting from $p$ toward $q$ (denoted as $\mathbf{h}_{p,e}^{(l)}$) is updated using a message-passing mechanism on the KG $\mathcal{G}$:
\begin{equation}
    \tilde{\mathbf{h}}_{p,e}^{(l)} = \text{ReLU} \left( \mathbf{W}^{(l)} \sum_{e' \in \mathcal{N}(e)} ( \mathbf{h}_{p,e'}^{(l-1)} \odot \hat{\mathbf{r}}_{e',e}^{(l)} ) \right),
\end{equation}
where $\mathcal{N}(e)$ denotes the neighbors of $e$ in $\mathcal{G}$. 

To mitigate numerical oversmoothing and structural feature homogenization, a gated-residual-flow mechanism is implemented, whereby the initial drug features are preserved throughout the propagation. The hidden state is updated via a gating weight $g^{(l)}$:
\begin{equation}
    g^{(l)} = \sigma\left( \mathbf{W}_{gate}^{(l)} [\tilde{\mathbf{h}}_{p,e}^{(l)} ; \mathbf{h}_{p,e}^{(0)}] \right),
\end{equation}
and
\begin{equation}
    \mathbf{h}_{p,e}^{(l)} = g^{(l)} \cdot \tilde{\mathbf{h}}_{p,e}^{(l)} + (1 - g^{(l)}) \cdot \mathbf{h}_{p,e}^{(0)},
\end{equation}
where $\mathbf{h}_{p,e}^{(0)} = \mathbf{f}_p$ is the initial drug feature from the previous module.

In contrast to conventional architectures that are restricted to final-layer representations, a multi-scale feature fusion strategy is adopted by CrossADR, whereby the latent embeddings of target drugs are extracted at each layer to capture the hierarchical evolution of knowledge flow. For the pair $(p, q)$, we obtain sequences of latent vectors $\mathbf{H}_p = \{\mathbf{h}_{p,q}^{(1)}, \dots, \mathbf{h}_{p,q}^{(L)}\}$ and $\mathbf{H}_q = \{\mathbf{h}_{q,p}^{(1)}, \dots, \mathbf{h}_{q,p}^{(L)}\}$. Based on the above architecture, in order to effectively fuse these multi-layer features, bi-directional cross-attention is then applied:
\begin{equation}
    \mathbf{A} = \text{Softmax} \left( \frac{(\mathbf{W}_{cross} \mathbf{H}_p) \mathbf{H}_q^T}{\sqrt{d}} \right),
\end{equation}
The cross-attended representations are calculated as $\hat{\mathbf{H}}_p = \mathbf{A} \mathbf{H}_q$ and $\hat{\mathbf{H}}_q = \mathbf{A}^T \mathbf{H}_p$. Finally, the output of the cross-layer GNN module $\mathbf{h}_{out1}^{p,q}$ is the concatenation of the flattened attended sequences:
\begin{equation}
    \mathbf{h}_{out1}^{p, q} = \left[ \text{vec}(\hat{\mathbf{H}}_p) ; \text{vec}(\hat{\mathbf{H}}_q) \right],
\end{equation}
where $\mathbf{h}_{out1}^{p, q} \in \mathbb{R}^{2 \cdot L \cdot d_1}$, capturing the complex inter-dependencies between drugs across different hop-distances in the KG.

\subsubsection{Cross-level attention module: Learnable space for ADR Embeddings}

The embedded vector $\mathbf{h}_{out1}^{p, q}$ captures information at the molecular level and network-based interactions. To integrate the latent organ-level ADR associations, CrossADR introduces a learnable embedding space for ADR labels. Unlike fixed categorical representations, this module maps drug-pair features into a multi-organ response space through a hierarchical attention mechanism.

Initially, $\mathbf{h}_{out1}^{p, q}$ is projected to generate preliminary ADR probability scores $\mathbf{S}_{1} \in \mathbb{R}^{15}$ for the 15 organ categories:
\begin{equation}
    \mathbf{S}_{1} = \sigma\left( \mathbf{W}_{rel1} (\mathbf{h}_{out1}^{p, q}) + \mathbf{b}_{rel1} \right),
\end{equation}
where $\mathbf{W}_{rel1} \in \mathbb{R}^{15 \times 2dL}$ and $\mathbf{b}_{rel1}$ are learnable parameters. 

To represent the presence or absence of ADRs in a continuous space, two learnable embedding sets are defined: the Positive ADR Space $\mathbf{E}_{+} \in \mathbb{R}^{15 \times d_2}$ and the Negative ADR Space $\mathbf{E}_{-} \in \mathbb{R}^{15 \times d_2}$, where $d_2$ is the dimension of the organ embeddings. For organ $i$, the score $\mathbf{S}_{1}(i)$ acts as a gating signal to fuse the previous latent representations, forming the Gated Embeddings $\mathbf{H}_{initial} \in \mathbb{R}^{15 \times d_2}$:
\begin{equation}
    g_i = \sigma(\mathbf{S}_{1}(i)),
\end{equation}
and
\begin{equation}
     \mathbf{H}_{initial}^{(i)} = g_i \cdot \mathbf{E}_{+}(i, :) + (1 - g_i) \cdot \mathbf{E}_{-}(i, :),
\end{equation}
where $\sigma$ denotes the sigmoid activation. This allows the model to capture the uncertainty and intensity of potential ADRs across different systems. Then, to model the complex co-occurrence and biological correlations between different organ systems, organ-level self-attention is applied on the initial gated features:
\begin{equation}
    \mathbf{H}_{attn} = \text{MultiHeadAttention}(\mathbf{H}_{initial}, \mathbf{H}_{initial}, \mathbf{H}_{initial}),
\end{equation}
\begin{equation}
    \mathbf{H}_{refined} = \tanh(\mathbf{H}_{initial} + \mathbf{H}_{attn}).
\end{equation}
The final organ-level feature vector $\mathbf{h}_{out2}^{p, q}$ is obtained by a weighted pooling based on the scores $\mathbf{S}_1$ and a global mean residual:
\begin{equation}
    \mathbf{w}_{pool} = \text{Softmax}(\mathbf{S}_1),
\end{equation}
and
\begin{equation}
     \mathbf{h}_{out2}^{p, q} = \sum_{i=1}^{15} (\mathbf{w}_{pool}(i) \cdot \mathbf{H}_{refined}(i, :)) + \text{Mean}(\mathbf{H}_{initial}).
\end{equation}

Finally, to effectively fuse the molecular-level network features $\mathbf{h}_{out1}^{p, q}$ with the organ-level latent features $\mathbf{h}_{out2}^{p, q}$, an organ-molecule cross-level cross-attention mechanism is applied.

First, the organ feature $\mathbf{h}_{out2}^{p, q}$ is projected into the molecular feature space using a transformation matrix $\mathbf{W}_t \in \mathbb{R}^{2dL \times d_2}$:
\begin{equation}
    \mathbf{h}_{out2}^{p, q\prime} = \mathbf{W}_t \mathbf{h}_{out2}^{p, q}.
\end{equation}

\begin{figure*}[!t]
\centering
\includegraphics[width=0.80\textwidth]{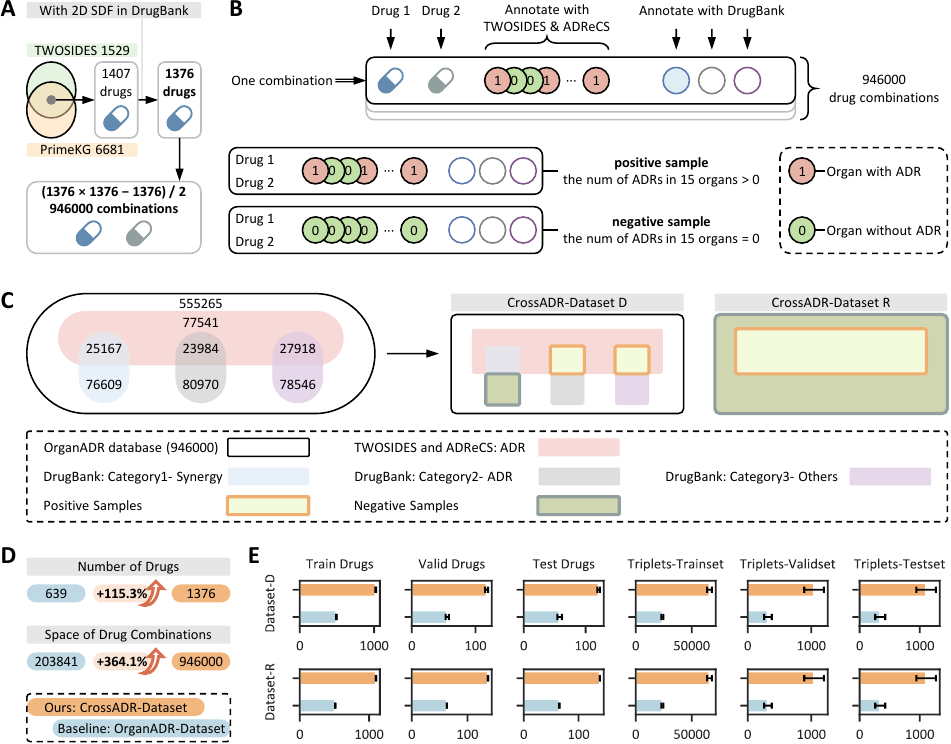}
\caption{Construction and statistics of the CrossADR-Dataset. (A) Determine the drugs to be analyzed for CrossADR. (B) Construction of one sample (one drug combination and its annotations); definition of positive sample and negative sample. (C) CrossADR database, CrossADR-Dataset D and CrossADR-Dataset R. (D, E) Comparison of the number of drugs, total combination space and other metrics between OrganADR-Dataset and the proposed CrossADR-Dataset.}\label{fig5}
\end{figure*}

The attention weights are then calculated by measuring the element-wise interaction between the molecular embedding and the transformed organ embedding, following the equations below:
\begin{equation}
    \mathbf{A}_{score} = \mathbf{h}_{out1}^{p, q} \odot \mathbf{h}_{out2}^{p, q\prime},
\end{equation}
and
\begin{equation}
    \mathbf{A}_{weight} = \text{Softmax}(\mathbf{A}_{score}).
\end{equation}
Cross-level weighted representation $\mathbf{h}_{out3}^{p, q} \in \mathbb{R}^{2dL}$ is computed as:
\begin{equation}
    \mathbf{h}_{out3}^{p, q} = \mathbf{A}_{weight} \odot \mathbf{h}_{out1}^{p, q}.
\end{equation}

Finally, to make the final prediction, the original molecular flow features, the global organ latent features, and the cross-attended features would be concatenated as $\mathbf{E} \in \mathbb{R}^{4dL + d_2}$:
\begin{equation}
    \mathbf{E} = [\mathbf{h}_{out1}^{p, q} ; \mathbf{h}_{out2}^{p, q} ; \mathbf{h}_{out3}^{p, q}].
\end{equation}

The final prediction scores $\mathbf{S} \in \mathbb{R}^{15}$ for ADRs in each organ or system are obtained through an output linear layer:
\begin{equation}
    \mathbf{S} = \sigma(\mathbf{W}_{output} \mathbf{E} + \mathbf{b}_{output}),
\end{equation}
where $\mathbf{W}_{output} \in \mathbb{R}^{15 \times (4dL + d_2)}$. Using a threshold of 0.5, the final binary ADR prediction $\hat{\mathcal{A}}_{p,q}$ is generated:
\begin{equation}
    \hat{a}^{i}_{p,q} = \begin{cases} 1 & \text{if } \mathbf{S}_i \geq 0.5 \\ 0 & \text{if } \mathbf{S}_i < 0.5 \end{cases}, \quad i = 1, 2, \ldots, 15.
\end{equation}

\subsection{Dataset: raw datasets and construction of CrossADR-Dataset}

\subsubsection{Raw datasets}
For the accurate prediction of organ-level adverse drug reactions (ADRs) within emerging combination therapies, TWOSIDES \cite{TWOSIDES} is utilized as the primary and most exhaustive database. Comprehensive prior knowledge is established through the biomedical knowledge graph, PrimeKG \cite{PrimeKG}. During the phase of data processing, ADReCS \cite{ADReCS2} is implemented to ensure that adverse drug reactions are mapped to organs. Furthermore, DrugBank \cite{DrugBank6} is employed for the annotation of drug combinations. Within this framework, DrugBank IDs (DBxxxxx) are systematically cross-referenced and matched to alternative identifiers to maintain data consistency.

\subsubsection{Construction of CrossADR-Dataset}
Following ref.\cite{OrganADR}, as shown in Fig.\ref{fig4} and Fig.\ref{fig5}, construction of CrossADR-Dataset includes two stages: (1) annotation of drug combinations; (2) construction of CrossADR-DataBase and CrossADR-Dataset D and CrossADR-Dataset R (Table.\ref{tab:dataset}).

\begin{table*}[!t]
\caption{\textbf{Statistic of the CrossADR-Dataset.} The symbols $|\mathcal{D}_{\text{train}}|$, $|\mathcal{D}_{\text{valid}}|$, and $|\mathcal{D}_{\text{test}}|$ in the main manuscript represent Train/Valid/Test drug sets (in the table header), respectively. The symbols $|\mathcal{C}_{\text{train}}|$, $|\mathcal{C}_{\text{valid}}|$, and $|\mathcal{C}_{\text{test}}|$ in the main manuscript represent the Triplets in the Trainset/Validset/Testset (in the table header). Each triplet refers to a single sample, with a balanced $1:1$ ratio between positive and negative samples. \label{tab:dataset}}
\tabcolsep=0pt
\begin{tabular*}{\textwidth}{@{\extracolsep{\fill}}lccccccc@{\extracolsep{\fill}}}
\toprule%
Type & Seed & Train Drugs & Valid Drugs & Test Drugs & Triplets-Trainset & Triplets-Validset & Triplets-Testset \\
\midrule
\multirow{10}{*}{CrossADR-Dataset D} & 10 & 1022 & 119 & 121 & 64190 & 1118 & 1240 \\
                    & 15 & 1017 & 120 & 122 & 64826 & 1160 & 1142 \\
                    & 20 & 1021 & 117 & 119 & 64224 & 1040 & 1234 \\
                    & 25 & 1023 & 119 & 123 & 65070 & 1260 & 1032 \\
                    & 30 & 1028 & 115 & 119 & 62922 & 1234 & 1318 \\
                    & 35 & 1020 & 120 & 122 & 68636 & 956  & 798  \\
                    & 40 & 1020 & 118 & 125 & 65282 & 1016 & 1264 \\
                    & 45 & 1021 & 116 & 119 & 66854 & 1010 & 1016 \\
                    & 50 & 1026 & 115 & 119 & 68672 & 834  & 970  \\
                    & 55 & 1027 & 114 & 122 & 68572 & 802  & 1046 \\
\midrule
\multirow{10}{*}{CrossADR-Dataset R} & 10 & 1100 & 137 & 139 & 64190 & 1118 & 1240 \\
                    & 15 & 1100 & 137 & 139 & 64826 & 1160 & 1142 \\
                    & 20 & 1100 & 137 & 139 & 64224 & 1040 & 1234 \\
                    & 25 & 1100 & 137 & 139 & 65070 & 1260 & 1032 \\
                    & 30 & 1100 & 137 & 139 & 62922 & 1234 & 1318 \\
                    & 35 & 1100 & 137 & 139 & 68636 & 956  & 798  \\
                    & 40 & 1100 & 137 & 139 & 65282 & 1016 & 1264 \\
                    & 45 & 1100 & 137 & 139 & 66854 & 1010 & 1016 \\
                    & 50 & 1100 & 137 & 139 & 68672 & 834  & 970  \\
                    & 55 & 1100 & 137 & 139 & 68572 & 802  & 1046 \\
\botrule
\end{tabular*}
\end{table*}

The CrossADR-Database is constructed through a systematic annotation process of drug pairs. During this procedure, both the ADR data and their respective sources are integrated. Initially, a pool of $1,376 \times 1,376$ possible combinations is generated based on the 1376 drugs selected for analysis. To refine this list, self-paired combinations (e.g., "drug $p$ - drug $p$") are excluded. Furthermore, the order of drugs within a pair is considered irrelevant, meaning "drug $p$ - drug $q$" and "drug $q$ - drug $p$" are treated as the same entity. Consequently, the total number of unique drug combinations is determined by the following calculation:
$ (1,376 \times 1,376 - 1,376) \times \frac{1}{2} = 946,000 $.

The ADR information for these pairs is then mapped into a binary vector of length 15. This vector represents the presence or absence of ADRs across 15 specific organs, with data retrieved from the TWOSIDES and ADReCS databases. Additionally, drug-drug interactions from DrugBank are classified into three distinct categories: ADR, synergy, and others. It is important to note that DrugBank data is utilized only for auxiliary sample selection and is not involved in the actual ADR annotation. Once these steps are completed, the OrganADR-Database is finalized. All drug pairs are then categorized based on their information sources and the intersections between them. It should be noted that the data sources for positive and negative samples are determined during the construction of OrganADR-Dataset D and OrganADR-Dataset R. 

In Dataset D, DrugBank records are utilized to identify negative samples. Positive samples in this dataset are defined as drug pairs recorded by TWOSIDES that are not classified as synergistic by DrugBank. Conversely, negative samples are composed of drug pairs explicitly recorded as synergistic within DrugBank. In Dataset R, a random selection method is applied to define negative samples. Here, positive samples are identified as the drug pairs documented in TWOSIDES. Negative samples are simply defined as those drug pairs that are not recorded by the TWOSIDES database.

The complete collection of positive and negative samples is not utilized for model training, validation, and testing directly. Instead, these samples function as supersets from which the specific training, validation, and testing partitions are derived, as shown below: 

Let the selected set of 1,376 drugs be represented by $\mathcal{V}$ (Fig. \ref{fig5}a). This set is partitioned into three mutually exclusive subsets:
\begin{equation}
    \mathcal{V} = \mathcal{V}_\text{train} \cup \mathcal{V}_\text{valid} \cup \mathcal{V}_\text{test} \quad \text{and} \quad \mathcal{V}_\text{train} \cap \mathcal{V}_\text{valid} \cap \mathcal{V}_\text{test} = \emptyset.
\end{equation}

Furthermore, the positive drug pair sets for training, validation, and testing are denoted by $\mathcal{C}_\text{p-train}$, $\mathcal{C}_\text{p-valid}$, and $\mathcal{C}_\text{p-test}$, respectively. Similarly, the negative sets are represented by $\mathcal{C}_\text{n-train}$, $\mathcal{C}_\text{n-valid}$, and $\mathcal{C}_\text{n-test}$. Herein, the initial positive and negative samples in CrossADR-Dataset D or R are denoted as $\mathcal{S}_\text{p}$ and $\mathcal{S}_\text{n}$.

\begin{itemize}
    \item The positive drug pairs (where at least one ADR is documented in TWOSIDES) are generated as follows:
    \begin{equation}
      \begin{cases}
          \mathcal{C}_\text{p-train} = \{(p, a_{p,q}^i, q): \ p \in \mathcal{V}_\text{train}, \ q \in \mathcal{V}_\text{train}, \ (p,q) \in \mathcal{S}_\text{p}\} \\
          \mathcal{C}_\text{p-valid} = \{(p, a_{p,q}^i, q): \ p \in \mathcal{V}_\text{valid}, \ q \in \mathcal{V}_\text{valid}, \ (p,q) \in \mathcal{S}_\text{p}\} \\
          \mathcal{C}_\text{p-test} = \{(p, a_{p,q}^i, q): \ p \in \mathcal{V}_\text{test}, \ q \in \mathcal{V}_\text{test}, \ (p,q) \in \mathcal{S}_\text{p}\}
      \end{cases},
    \end{equation}
    \item The negative drug pairs (where no ADR is documented in TWOSIDES) are generated as follows:
    \begin{equation}
      \begin{cases}
          \mathcal{C}_\text{n-train} = \{(p, a_{p,q}^i, q): \ p \in \mathcal{V}_\text{train}, \ q \in \mathcal{V}_\text{train}, \ (p,q) \in \mathcal{S}_\text{n}\} \\
          \mathcal{C}_\text{n-valid} = \{(p, a_{p,q}^i, q): \ p \in \mathcal{V}_\text{valid}, \ q \in \mathcal{V}_\text{valid}, \ (p,q) \in \mathcal{S}_\text{n}\} \\
          \mathcal{C}_\text{n-test} = \{(p, a_{p,q}^i, q): \ p \in \mathcal{V}_\text{test}, \ q \in \mathcal{V}_\text{test}, \ (p,q) \in \mathcal{S}_\text{n}\}
      \end{cases},
    \end{equation}
\end{itemize}

To maintain data balance, the quantity of positive drug pairs is matched with the negative drug pairs across all sets. This procedure is performed in accordance with refs.\cite{OrganADR, EmerGNN, SumGNN}. Statistics for the CrossADR-Dataset under random seeds are shown in Table \ref{tab:dataset}.

\subsection{Experiment: train and evaluate CrossADR}

\subsubsection{Loss function}
Since multiple interactions between two drugs may occur, the binary cross-entropy loss is applied to the model. The loss is formulated as:
\begin{equation}
    \mathcal{L} = - \frac{1}{15} \sum_{i=1}^{15} \left[ a^i_{p,q} \log(\mathbf{S}_i) + (1 - a^i_{p,q}) \log(1 - \mathbf{S}_i) \right],
\end{equation}
where \( a^i_{p,q} \) represents the actual binary label for the \(i\)-th ADR, \( \mathbf{S}_i \) represents the predicted probability of the \(i\)-th ADR.

\subsubsection{Optimizer}
The Adam optimizer \cite{Adam} is employed. Model parameters are optimized by minimizing the defined loss function.

\subsubsection{Evaluation metrics}
ROC-AUC and PR-AUC are utilized as the primary evaluation metrics. These metrics are chosen because sensitivity to the classification threshold is avoided. 
\begin{itemize}
    \item \textbf{ROC-AUC}: The area under the curve is calculated for TPR vs. FPR across various threshold settings.
    \vspace{4pt}
    \item \textbf{PR-AUC}: The area under the curve is determined for Precision vs. Recall at different thresholds.
\end{itemize}
The constituent components are defined as:
\begin{equation}
    \text{TPR} = \frac{\text{TP}}{\text{TP} + \text{FN}}, \quad \text{FPR} = \frac{\text{FP}}{\text{FP} + \text{TN}}, 
\end{equation}
and
\begin{equation}
    \text{Precision} = \frac{\text{TP}}{\text{TP} + \text{FP}}, \quad \text{Recall} = \frac{\text{TP}}{\text{TP} + \text{FN}}. 
\end{equation}

For a comprehensive assessment of the classification of positive and negative samples, Accuracy, Precision, Recall, and Hamming loss are integrated as supplementary metrics.
\begin{itemize}
    \item \textbf{Accuracy}: The overall correctness. It is defined as the ratio of correctly predicted instances to the total number of samples.
\end{itemize}
\begin{equation}
    \text{Accuracy} = \frac{\text{TP} + \text{TN}}{\text{TP} + \text{TN} + \text{FP} + \text{FN}}
\end{equation}

\begin{itemize}
    \item \textbf{Precision}: High importance is placed on precision when false positive costs are significant. The accuracy of positive predictions:
\end{itemize}
\begin{equation}
    \text{Precision} = \frac{\text{TP}}{\text{TP} + \text{FP}}
\end{equation}

\begin{itemize}
    \item \textbf{Recall}: The model's ability to identify all relevant instances. This metric is prioritized when false negatives are costly.
\end{itemize}
\begin{equation}
    \text{Recall} = \frac{\text{TP}}{\text{TP} + \text{FN}}
\end{equation}

In these equations, TP denotes correctly identified positive cases, and TN represents correctly identified negative cases. Conversely, negative cases incorrectly labeled as positive are denoted by FP, while positive cases incorrectly identified as negative are represented by FN.

\begin{itemize}
    \item \textbf{F1-Score}: A balance between precision and recall is maintained by the F1-Score. It is calculated as the harmonic mean of the two and is frequently applied to imbalanced datasets.
\end{itemize}
\begin{equation}
    \text{F1-Score} = 2 \times \frac{\text{Precision} \cdot \text{Recall}}{\text{Precision} + \text{Recall}}
\end{equation}

\begin{itemize}
    \item \textbf{Hamming loss}: The fraction of incorrect labels in the multi-label task is measured by Hamming loss. The average number of misclassified labels is represented.
\end{itemize}
\begin{equation}
    \text{Hamming loss} = \frac{1}{N \times L} \sum_{i=1}^{N} \sum_{j=1}^{L} \mathbb{I}(y_{ij} \ne \hat{y}_{ij})
\end{equation}
where $N$ is defined as the total number of samples, and $L$ is denoted as the number of labels assigned to each sample. The actual $j$-th label of the $i$-th sample is represented by $y_{ij}$, while the predicted label is represented by $\hat{y}_{ij}$. An indicator function $\mathbb{I}(y_{ij} \ne \hat{y}_{ij})$ is employed; 1 is assigned if $y_{ij} \ne \hat{y}_{ij}$, and 0 is assigned otherwise.

\section{Results}\label{sec3}

\subsection{CrossADR: organ-level adverse drug reaction prediction}

Predicting ADRs for combination pharmacotherapy is a critical yet challenging task, especially for emerging drugs where historical interaction data is sparse. As illustrated in Fig.\ref{fig1}A, the proposed CrossADR framework addresses this by formulating the problem into a multi-label classification task across 15 organs. By taking drug pairs with their molecular features as input, CrossADR maps molecular and network-level interactions to clinical ADR outcomes at the organ level. The comprehensive workflow, shown in Fig.\ref{fig1}B-C, spans from initial drug feature extraction to the prediction of ADRs, facilitating downstream biological insights such as Protein-Protein Interaction (PPI) network mapping and pathway enrichment analysis.

\subsection{Architecture of CrossADR}

The CrossADR framework is designed as a hierarchical information processing pipeline that bridges the gap between microscopic drug properties and macroscopic clinical outcomes. As illustrated in Fig.\ref{fig1}C and Fig.\ref{fig2}, the architecture is characterized by two core innovations that distinguish it from previous GNN-KG based models like OrganADR or EmerGNN: a cross-layer feature integration strategy and a cross-level associative learning mechanism.

\begin{itemize}
    \item \textbf{Cross-layer Feature Integration:} Unlike conventional GNN architectures that rely solely on the final hidden state, CrossADR implements a gated-residual-flow GNN. In CrossADR, this module (Fig.\ref{fig2}B) utilizes drug-dependent attention scores to weight the Knowledge Graph (KG) edges dynamically. By adopting a multi-scale fusion strategy, the model captures the hierarchical evolution of knowledge flow across different hop-distances in the KG. Bi-directional cross-attention is then applied to these multi-layer sequences, ensuring that the structural context of both drugs is integrated at every level of abstraction.
    
    \item \textbf{Cross-level Associative Learning:} Importantly, as shown in Fig.\ref{fig2}E, unlike existing methods that necessitate a pre-defined label association matrix or fixed hierarchical knowledge of ADRs, CrossADR’s embedding space is entirely learnable and data-driven. This approach circumvents the limitations of incomplete or biased prior medical knowledge. By allowing the model to autonomously capture latent dependencies between organ-level responses, CrossADR can identify non-obvious clinical associations that are often overlooked by rigid, manually-curated correlation structures, thereby enhancing the robustness of multi-label ADR predictions in complex pharmacotherapy scenarios.
\end{itemize}

Finally, a cross-level cross-attention module integrates these molecular-level network with the refined organ-level latent features. This dual-pathway design—processing "bottom-up" molecular information and "top-down" biological associations simultaneously—enables CrossADR to achieve superior predictive accuracy and meanwhile provides a more interpretable representation of how drug interactions manifest as specific organ-level ADRs.

\subsection{Generation of different KGs}

\begin{table*}[!t]
\caption{\textbf{Overview of the KGs' edges used in this manuscript.} The semantic relations, including source and target node categories, and the total edge count for each triplet type are shown. The columns "KG: Basic" and "KG: Ablation 1–3" define the corresponding edge subsets for the 4 KGs.\label{tab:kg}}
\tabcolsep=0pt
\begin{tabular*}{\textwidth}{@{\extracolsep{\fill}}lllccccc}
\toprule%
Relation & Type of Source Node & Type of Target Node & Number of Edges & KG: Basic & KG: Ablation 1 & KG: Ablation 2 & KG: Ablation 3 \\
\midrule
ppi & gene/protein & gene/protein & 642,150 & $\checkmark$ & $\checkmark$ & & $\checkmark$ \\
associated with & effect/phenotype & gene/protein & 3,330 & $\checkmark$ & $\checkmark$ & & \\
associated with & gene/protein & effect/phenotype & 3,330 & $\checkmark$ & $\checkmark$ & & \\
parent-child & effect/phenotype & effect/phenotype & 37,472 & $\checkmark$ & & & \\
target & drug & gene/protein & 16,380 & $\checkmark$ & $\checkmark$ & & $\checkmark$ \\
target & gene/protein & drug & 16,380 & $\checkmark$ & $\checkmark$ & & $\checkmark$ \\
enzyme & drug & gene/protein & 5,317 & $\checkmark$ & $\checkmark$ & & $\checkmark$ \\
enzyme & gene/protein & drug & 5,317 & $\checkmark$ & $\checkmark$ & & $\checkmark$ \\
transporter & drug & gene/protein & 3,092 & $\checkmark$ & $\checkmark$ & & $\checkmark$ \\
transporter & gene/protein & drug & 3,092 & $\checkmark$ & $\checkmark$ & & $\checkmark$ \\
carrier & drug & gene/protein & 864 & $\checkmark$ & $\checkmark$ & & $\checkmark$ \\
carrier & gene/protein & drug & 864 & $\checkmark$ & $\checkmark$ & & $\checkmark$ \\
side effect & drug & effect/phenotype & 64,784 & $\checkmark$ & $\checkmark$ & $\checkmark$ & \\
side effect & effect/phenotype & drug & 64,784 & $\checkmark$ & $\checkmark$ & $\checkmark$ & \\
associated with & disease & gene/protein & 80,411 & $\checkmark$ & & & $\checkmark$ \\
associated with & gene/protein & disease & 80,411 & $\checkmark$ & & & $\checkmark$ \\
phenotype present & disease & effect/phenotype & 150,317 & $\checkmark$ & & $\checkmark$ & \\
phenotype present & effect/phenotype & disease & 150,317 & $\checkmark$ & & $\checkmark$ & \\
phenotype absent & disease & effect/phenotype & 1,193 & $\checkmark$ & & $\checkmark$ & \\
phenotype absent & effect/phenotype & disease & 1,193 & $\checkmark$ & & $\checkmark$ & \\
contraindication & disease & drug & 30,675 & $\checkmark$ & & $\checkmark$ & $\checkmark$ \\
contraindication & drug & disease & 30,675 & $\checkmark$ & & $\checkmark$ & $\checkmark$ \\
indication & disease & drug & 9,388 & $\checkmark$ & & $\checkmark$ & $\checkmark$ \\
indication & drug & disease & 9,388 & $\checkmark$ & & $\checkmark$ & $\checkmark$ \\
off-label use & disease & drug & 2,568 & $\checkmark$ & & $\checkmark$ & $\checkmark$ \\
off-label use & drug & disease & 2,568 & $\checkmark$ & & $\checkmark$ & $\checkmark$ \\
parent-child & disease & disease & 64,388 & $\checkmark$ & & $\checkmark$ & $\checkmark$ \\
\botrule
\end{tabular*}
\begin{tablenotes}%
\end{tablenotes}
\end{table*}

In the architecture of Graph Neural Network (GNN)-based models, the KG is recognized as an indispensable module for capturing high-order biological relationships. As information carriers, the richness and diversity of KGs are crucial for the model's representative ability. Although the focus of this study is the development of the CrossADR model's architecture, different KGs are strategically constructed to comprehensively evaluate the performance and robustness of the proposed framework under varying information densities.

As illustrated in Fig.\ref{fig3}A, the foundational "KG: Basic" is constructed based on PrimeKG. Following the preprocessing protocols established in previous studies \cite{EmerGNN, OrganADR}, specific node types and relations associated with drug mechanisms are prioritized. The basic KG comprises four primary entity types: drug, gene/protein, effect/phenotype, and disease. As detailed in Table \ref{tab:kg}, a comprehensive network is formed with 27 types of semantic relations. Notably, Protein-Protein Interactions (PPIs) constitute the largest portion of the graph with 642,150 edges, while drug-related interactions such as "target", "enzyme", and "side effect" are integrated to provide a multi-faceted pharmacological context. Synergistic interaction edges between drugs are explicitly removed to prevent any potential knowledge leakage. In addition, 15 types of ADRs (15 organs) connecting drugs, as well as node-to-node self-loops, are also included in the final KG used for training.

Furthermore, three ablated KGs are derived from the basic version to facilitate downstream performance assessment and sensitivity analysis (Fig.\ref{fig3}B-D). These KGs are summarized as follows:
\begin{itemize}
    \item \textbf{KG: Ablation 1} is designed to focus on the immediate biological neighborhood of drugs. In this version, all disease-related nodes and their associated edges (e.g., "contraindication") are removed, and only drug, protein, and phenotype nodes are retained.
    \item \textbf{KG: Ablation 2} is constructed to evaluate the model's performance when protein-level information is completely absent. All gene/protein nodes and their corresponding interactions, including PPIs and drug-target relations, are excluded, leaving only the clinical and phenotypic associations.
    \item \textbf{KG: Ablation 3} represents a configuration where phenotypic information is removed. All "effect/phenotype" nodes, "side effect" relations and other related edges are purged, focusing instead on the relationships between diseases, proteins and drugs.
\end{itemize}

These ablated KGs serve as the topological foundation for the subsequent comparative experiments, allowing for a granular understanding of how different biological knowledge components contribute to or affect on the prediction of organ-level ADRs.

\subsection{Construction and statistics of CrossADR-Dataset}

As shown in Fig.\ref{fig4} and Fig.\ref{fig5}, the CrossADR-Dataset is constructed through a pipeline involving drug selection and multi-source annotation. The construction of CrossADR-Dataset expands the analysis to a pool of 1,376 clinically relevant drugs. As shown in Fig.\ref{fig5}A, this results in a vast combination space of 946,000 unique drug-drug pairs. To ensure the chemical and biological representativeness of these drugs, their identifiers are cross-referenced across PrimeKG \cite{PrimeKG}, TWOSIDES \cite{TWOSIDES}, and DrugBank \cite{DrugBank6}, ensuring that all analyzed drugs possess both structural descriptors (SDFs) and network-based positions in the knowledge graph.

The annotation process (Fig.\ref{fig4}A) maps raw ADR symptoms from TWOSIDES to 15 specific organ systems using the ADReCS \cite{ADReCS2} hierarchy. Furthermore, we integrated DrugBank's interaction records as auxiliary information to distinguish between ADR-related interactions and synergistic effects (Fig.\ref{fig4}B). Following the methodology of OrganADR \cite{OrganADR}, two distinct benchmark datasets are established to evaluate the model under different scenarios:
\begin{itemize}
    \item \textbf{CrossADR-Dataset D}: This dataset utilizes DrugBank's synergistic records as a high-confidence source for negative samples, while positive samples are pairs documented in TWOSIDES that are not synergistic.
    \item \textbf{CrossADR-Dataset R}: In this more challenging configuration, negative samples are randomly selected from the complement set of documented ADR pairs in TWOSIDES, reflecting a broader pharmacological space with uncertain annotations.
\end{itemize}

Statistical comparison (Fig.\ref{fig5}D-E) highlights the expansion of CrossADR-Dataset. While the previous OrganADR-Dataset utilized approximately 500-600 drugs for training, the CrossADR-Dataset increases the number of training drugs to over 1,000 (e.g., an average of $1022.5 \pm 3.5$ for Dataset D across different random seeds and $1100$ for Dataset R). This expansion is further reflected in the triplet counts; for instance, the number of triplets in the training set increased from approximately 23,749 in OrganADR to an average of $65,924.8 \pm 2109.2$ drug pairs across seed in the CrossADR-Dataset.

Drug-disjoint partitioning (8:1:1 ratio).are performed. As detailed in Table \ref{tab:dataset}, the drugs are divided into $\mathcal{V}_{\text{train}}$, $\mathcal{V}_{\text{valid}}$, and $\mathcal{V}_{\text{test}}$, ensuring that drugs in the test set never appeared during training.

\begin{figure*}[!t]
\centering
\includegraphics[width=0.80\textwidth]{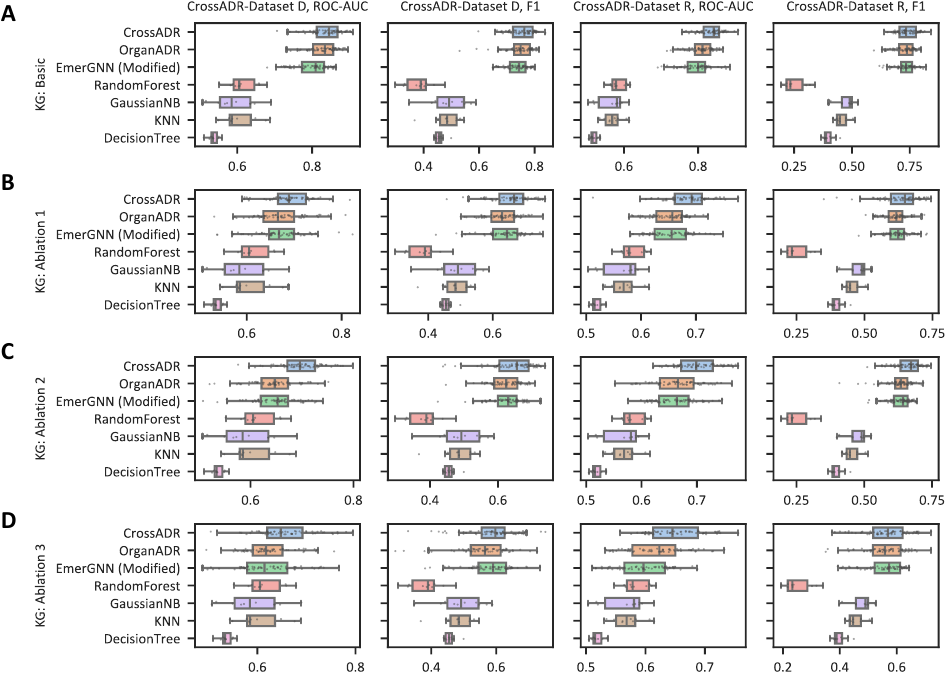}
\caption{Performance comparison of CrossADR against baseline models. The figure is organized into a $4 \times 4$ panel matrix: rows 1 to 4 correspond to different Knowledge Graph configurations (KG: Basic, KG: Ablation 1, KG: Ablation 2, and KG: Ablation 3, respectively); the first two columns present results for CrossADR-Dataset D (ROC-AUC and F1-score), while the last two columns show results for CrossADR-Dataset R (ROC-AUC and F1-score). In each boxplot, individual scatter points represent results from independent experimental runs, the central horizontal line indicates the median value, the box boundaries denote the upper and lower quartiles, and the whiskers extend to illustrate the overall data range excluding outliers. For each model in each subplot, N = 60.}\label{fig6}
\end{figure*}

\begin{table*}[!t]
\caption{\textbf{Performance of CrossADR with baseline models across 5 metrics.} CrossADR, OrganADR and EmerGNN are run three times with random seeds 10, 20, 30. Four ML models are run one time with fixed random state 1. The process above is repeated on ten datasets, which are generated with the random seeds 10, 15, 20, 25, 30, 35, 40, 45, 50 and 55. Given that the ratio of validation set to test set is 1:1, the two sets serve as mutual validation and test sets, and are tested twice. The reported values in table are the mean value and standard deviation calculated from a total of 60 runs for deep-learning models and 10 runs for machine-learning models. All evaluation metrics are shown as percentages (\%), where higher values are better for ROC-AUC, PR-AUC, ACC and F1, and lower values indicate better performance for HL. Best results in \textbf{bold}, second best \underline{underlined}. \label{tab:main_performance}}
\tabcolsep=0pt
\begin{tabular*}{\textwidth}{@{\extracolsep{\fill}}llccccc@{\extracolsep{\fill}}}
\toprule%
Datasource & Model & PR-AUC & ROC-AUC & ACC & F1 & HL \\
\midrule
\multirow{7}{*}{CrossADR-Dataset D} 
 & CrossADR & \textbf{83.83\% $\pm$ 3.99\%} & \textbf{83.75\% $\pm$ 4.31\%} & \textbf{76.44\% $\pm$ 3.91\%} & \textbf{75.32\% $\pm$ 4.76\%} & \textbf{24.25\% $\pm$ 3.93\%} \\
 & OrganADR & \underline{83.17\% $\pm$ 4.25\%} & \underline{82.76\% $\pm$ 4.04\%} & \underline{75.34\% $\pm$ 3.87\%} & \underline{74.52\% $\pm$ 5.32\%} & \underline{24.66\% $\pm$ 3.87\%} \\
 & EmerGNN (Modified) & 81.57\% $\pm$ 4.32\% & 79.86\% $\pm$ 4.51\% & 72.71\% $\pm$ 4.68\% & 73.69\% $\pm$ 4.13\% & 27.29\% $\pm$ 4.68\% \\
 & RandomForest & 61.94\% $\pm$ 4.68\% & 61.31\% $\pm$ 3.98\% & 56.14\% $\pm$ 2.59\% & 38.37\% $\pm$ 5.35\% & 43.86\% $\pm$ 2.59\% \\
 & KNN & 58.20\% $\pm$ 4.05\% & 60.35\% $\pm$ 5.03\% & 56.91\% $\pm$ 3.04\% & 48.28\% $\pm$ 5.23\% & 43.09\% $\pm$ 3.04\% \\
 & GaussianNB & 57.76\% $\pm$ 5.26\% & 59.14\% $\pm$ 5.96\% & 56.98\% $\pm$ 5.19\% & 49.20\% $\pm$ 7.34\% & 43.02\% $\pm$ 5.19\% \\
 & DecisionTree & 52.10\% $\pm$ 0.84\% & 53.59\% $\pm$ 1.36\% & 53.58\% $\pm$ 1.36\% & 45.73\% $\pm$ 1.82\% & 46.42\% $\pm$ 1.36\% \\
\midrule
\multirow{7}{*}{CrossADR-Dataset R} 
 & CrossADR & \textbf{82.91\% $\pm$ 3.33\%} & \textbf{83.57\% $\pm$ 3.26\%} & \textbf{75.45\% $\pm$ 3.15\%} & \textbf{74.06\% $\pm$ 4.40\%} & \textbf{25.24\% $\pm$ 3.12\%} \\
 & OrganADR & \underline{79.79\% $\pm$ 3.48\%} & \underline{80.67\% $\pm$ 3.31\%} & \underline{73.25\% $\pm$ 3.17\%} & 72.76\% $\pm$ 4.55\% & \underline{26.75\% $\pm$ 3.17\%} \\
 & EmerGNN (Modified) & 79.26\% $\pm$ 3.82\% & 79.42\% $\pm$ 4.09\% & 72.61\% $\pm$ 3.80\% & \underline{73.16\% $\pm$ 4.27\%} & 27.39\% $\pm$ 3.80\% \\
 & RandomForest & 58.91\% $\pm$ 3.51\% & 58.47\% $\pm$ 2.32\% & 53.71\% $\pm$ 1.70\% & 24.86\% $\pm$ 5.22\% & 46.29\% $\pm$ 1.70\% \\
 & GaussianNB & 55.42\% $\pm$ 2.87\% & 56.59\% $\pm$ 3.81\% & 55.22\% $\pm$ 2.49\% & 47.48\% $\pm$ 4.44\% & 44.78\% $\pm$ 2.49\% \\
 & KNN & 54.96\% $\pm$ 2.03\% & 56.95\% $\pm$ 2.82\% & 54.31\% $\pm$ 1.92\% & 45.68\% $\pm$ 3.33\% & 45.69\% $\pm$ 1.92\% \\
 & DecisionTree & 51.12\% $\pm$ 0.61\% & 51.92\% $\pm$ 1.08\% & 51.91\% $\pm$ 1.08\% & 39.83\% $\pm$ 2.51\% & 48.09\% $\pm$ 1.08\% \\
\botrule
\end{tabular*}
\begin{tablenotes}
\end{tablenotes}
\end{table*}

\begin{table*}[!t]
\caption{\textbf{Significance test of CrossADR with baseline model OrganADR and EmerGNN.} Consistent with the previous annotation, the reported values in table are calculated from a total of 60 runs. The values reported for Model 1 and Model 2 represent the mean ROC-AUC across experimental runs. ** ($P < 0.01$), *** ($P < 0.001$). Note that EmerGNN implemented in this comparison is the Modified EmerGNN adapted in this manuscript.\label{tab:sig_1}}
\tabcolsep=0pt
\begin{tabular*}{\textwidth}{@{\extracolsep{\fill}}llcccccccccc@{\extracolsep{\fill}}}
\toprule%
& & \multicolumn{5}{@{}c@{}}{CrossADR (Model 1) vs OrganADR (Model 2)} & \multicolumn{5}{@{}c@{}}{CrossADR (Model 1) vs EmerGNN (Model 2)} \\
\cmidrule{3-7}\cmidrule{8-12}
Dataset & KG & Model 1 & Model 2 & P-Value & Sig & Effect Size & Model 1 & Model 2 & P-Value & Sig & Effect Size \\
\midrule
\multirow{4}{*}{CrossADR Dataset-D} 
& Basic & 83.7\% & 82.8\% & $2.24\times10^{-3}$ & ** & 0.413 & 83.7\% & 79.9\% & $8.96\times10^{-15}$ & *** & 1.327 \\
& Ablation 1 & 68.8\% & 66.9\% & $2.74\times10^{-3}$ & ** & 0.404 & 68.8\% & 66.9\% & $1.91\times10^{-3}$ & ** & 0.420 \\
& Ablation 2 & 69.6\% & 64.7\% & $1.18\times10^{-11}$ & *** & 1.084 & 69.6\% & 64.8\% & $8.99\times10^{-11}$ & *** & 1.016 \\
& Ablation 3 & 65.5\% & 62.4\% & $1.11\times10^{-4}$ & *** & 0.535 & 65.5\% & 62.2\% & $6.01\times10^{-4}$ & *** & 0.468 \\
\midrule
\multirow{4}{*}{CrossADR Dataset-R} 
&  Basic & 83.6\% & 80.7\% & $9.75\times10^{-11}$ & *** & 1.690 & 83.6\% & 79.4\% & $1.22\times10^{-18}$ & *** & 1.650 \\
& Ablation 1 & 68.4\% & 65.2\% & $1.21\times10^{-8}$ & *** & 0.734 & 68.4\% & 65.4\% & $3.50\times10^{-7}$ & *** & 0.651 \\
& Ablation 2 & 70.2\% & 66.4\% & $1.65\times10^{-10}$ & *** & 0.996 & 70.2\% & 66.0\% & $9.75\times10^{-11}$ & *** & 1.435 \\
& Ablation 3 & 65.1\% & 61.9\% & $4.49\times10^{-6}$ & *** & 0.652 & 65.1\% & 59.7\% & $1.45\times10^{-12}$ & *** & 1.154 \\
\botrule
\end{tabular*}
\end{table*}

\subsection{Comparison of CrossADR with baseline models}

To evaluate the predictive performance of the proposed CrossADR framework, a comprehensive comparative analysis is conducted across two benchmark datasets (CrossADR-Dataset D and Dataset R) under four distinct Knowledge Graph (KG) configurations. The performance of CrossADR is compared against six baseline models, including two state-of-the-art deep learning architectures specifically designed for drug-drug interaction (DDI) tasks—OrganADR \cite{OrganADR} and a modified version of EmerGNN \cite{EmerGNN}—and four traditional machine learning (ML) models: Random Forest (RF), K-Nearest Neighbors (KNN), Gaussian Naive Bayes (GNB), and Decision Tree (DT).

As summarized in Table \ref{tab:main_performance}, the overall performance on the foundational "KG: Basic" demonstrates that CrossADR consistently outperforms all baseline models across all five evaluation metrics. In CrossADR-Dataset D, CrossADR achieves a ROC-AUC of $83.75\% \pm 4.31\%$ and an F1-score of $75.32\% \pm 4.76\%$, representing a significant improvement over the second-best model, OrganADR (ROC-AUC: $82.76\% \pm 4.04\%$). A similar trend is observed in the more challenging CrossADR-Dataset R, where CrossADR maintains a high ROC-AUC of $83.57\% \pm 3.26\%$. Notably, the deep learning-based models (CrossADR, OrganADR, and EmerGNN) exhibit a substantial performance gap compared to traditional ML methods. For instance, the ROC-AUC of the best-performing ML model (RF) is approximately $61.31\%$ in Dataset D, which is $22.44\%$ lower than that of CrossADR, highlighting the necessity of integrating multi-scale biological knowledge via GNNs and the proposed architecture.

The robustness of CrossADR is further validated through significance testing across all four KG topologies, as detailed in Table \ref{tab:sig_1}. When compared to OrganADR and EmerGNN, CrossADR demonstrates statistically significant improvements in nearly every scenario. On the Basic KG, the $P$-values for ROC-AUC compared to OrganADR are $2.24 \times 10^{-3}$ and $9.75 \times 10^{-11}$ for Dataset D and R, respectively. Even under information-sparse conditions (Ablations 1--3), where performance generally declines for all models due to the removal of disease, protein, or phenotype nodes, CrossADR remains the superior model. For example, in KG: Ablation 2 (where protein information is removed), CrossADR achieves a mean ROC-AUC of $69.6\%$ (Dataset D) and $70.2\%$ (Dataset R), outperforming OrganADR by $4.9\%$ and $3.8\%$, respectively, with high significance ($P < 0.001$). The effect sizes (Cohen's d) are notably high, reaching $1.690$ in the Dataset R Basic KG scenario, indicating that the architectural innovations of CrossADR—specifically the cross-layer integration and learnable ADR embeddings—provide substantial gains that are not dependent on specific KG densities.

The comprehensive performance distribution is visually illustrated in Fig. \ref{fig6}, which presents the results of 60 independent experimental runs for the deep learning models and 10 runs for the ML models. The boxplots reveal that CrossADR not only achieves higher median values but also maintains relatively narrow interquartile ranges, suggesting high stability across different data partitions. In Fig. \ref{fig6}A (KG: Basic), the ROC-AUC and F1-score distributions for CrossADR are shifted significantly upward compared to EmerGNN and the ML baselines. As the KG complexity decreases from Fig. \ref{fig6}B to \ref{fig6}D, while the absolute performance of all models is impacted by the loss of topological information, CrossADR consistently maintains its lead. Specifically, in KG: Ablation 3 (Fig. \ref{fig6}D), where phenotypic information is absent, the median F1-score of CrossADR in Dataset D ($59.75\%$) remains higher than that of OrganADR ($56.54\%$) and EmerGNN ($59.06\%$), demonstrating the CrossADR's ability to effectively leverage cross-level information through its gated-residual-flow mechanism and associative learning architectures.

\begin{figure*}[!t]
\centering
\includegraphics[width=0.80\textwidth]{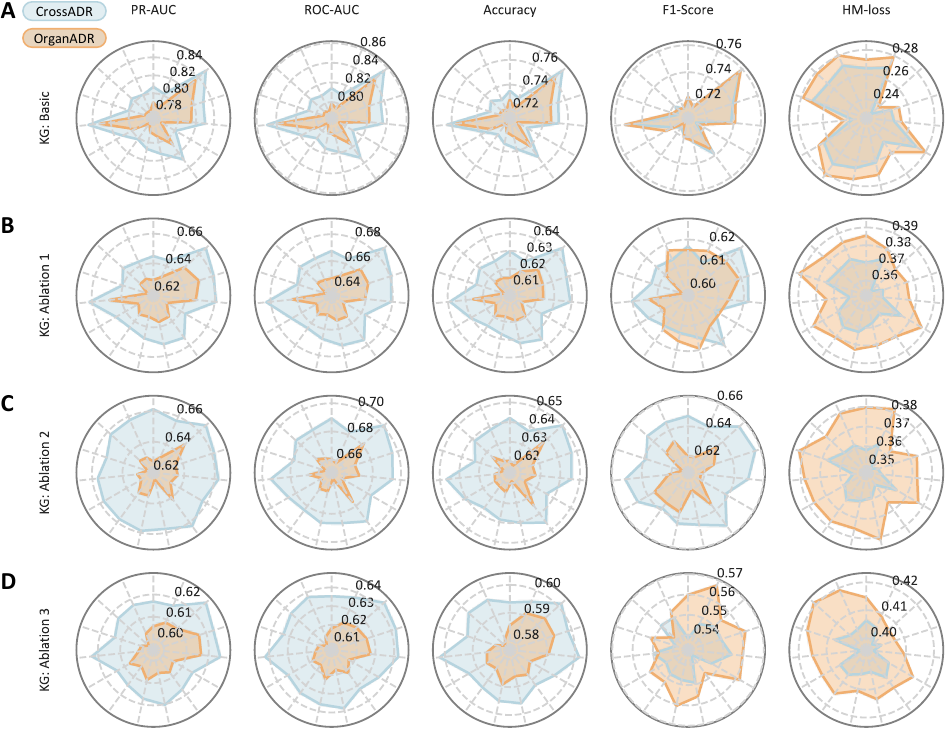}
\caption{Evaluation of CrossADR verses OrganADR across 15 organs. (A-D) Comprehensive evaluation across four KG configurations (Basic and Ablations 1--3) using five metrics: PR-AUC, ROC-AUC, Accuracy, F1-Score, and Hamming Loss (HM-loss). Each of the 15 axes represents a specific organ category.}\label{fig7}
\end{figure*}

\subsection{CrossADR's performance across 15 organs}

To further evaluate the fine-grained predictive capability of the proposed model, the performance of CrossADR is assessed across 15 specific organ categories. This granular analysis is conducted to verify whether the architectural advantages of CrossADR translate into consistent improvements across diverse physiological systems. As illustrated in Fig.\ref{fig7}, the performance of CrossADR is compared against the previously state-of-the-art OrganADR model using five metrics (PR-AUC, ROC-AUC, Accuracy, F1-Score, and Hamming Loss) under four distinct Knowledge Graph (KG) configurations. In each radar chart, the 15 axes represent individual organ systems, providing a multi-dimensional visualization of the models' robustness.

As shown in Fig.\ref{fig7}A, under the ``KG: Basic", CrossADR demonstrates a consistent lead over OrganADR across all 15 organs. For the PR-AUC metric, CrossADR achieves values ranging from $80.07\%$ to $85.08\%$, whereas OrganADR yields lower values between $77.96\%$ and $83.53\%$. Specifically, in Organ 3, CrossADR reaches a peak PR-AUC of $85.08\%$, surpassing OrganADR’s $83.53\%$. A similar trend is observed in ROC-AUC, where CrossADR maintains a range of $80.82\%$ to $85.54\%$, consistently outperforming OrganADR ($79.02\%$ to $84.30\%$). Regarding prediction accuracy, CrossADR records a minimum of $72.87\%$ (Organ 6) and a maximum of $76.99\%$ (Organ 3), while the accuracy of OrganADR remains consistently lower, with a range of $71.81\%$ to $76.21\%$. Notably, the HM-loss of CrossADR is reduced. For instance, in Organ 3, CrossADR achieves the lowest HM-loss of $23.01\%$, whereas OrganADR records $23.79\%$.

The superiority of CrossADR is maintained even under information-sparse conditions provided by the ablated KGs. As illustrated in Fig.\ref{fig7}B--D, although the absolute performance of both models declines when biological nodes (diseases, proteins, or phenotypes) are removed, CrossADR consistently encompasses a larger area in the radar charts than OrganADR. In the "KG: Ablation 2" scenario (Fig.\ref{fig7}C), where protein information is absent, CrossADR still achieves an ROC-AUC of $70.08\%$ in Organ 3, which is $1.94\%$ higher than the $68.14\%$ achieved by OrganADR. Furthermore, in the "KG: Ablation 3" configuration (Fig.\ref{fig7}D), where phenotypic information is removed, CrossADR maintains an F1-score between $53.58\%$ and $55.41\%$, whereas OrganADR's performance fluctuates between $54.19\%$ and $56.74\%$ with higher instability across organs.

\subsection{Ablation study}

\begin{table*}[!t]
\caption{\textbf{Significance test of CrossADR with two ablation studies.} Consistent with the previous annotation, the reported values in table are calculated from a total of 60 runs. The values reported for Model 1 and Model 2 represent the mean ROC-AUC across experimental runs. * ($P < 0.05$), ** ($P < 0.01$), *** ($P < 0.001$). Ablation 1 and Ablation 2 refer to ablated variants of CrossADR, with detailed information in the main text.\label{tab:sig_2}}
\tabcolsep=0pt
\begin{tabular*}{\textwidth}{@{\extracolsep{\fill}}llcccccccccc@{\extracolsep{\fill}}}
\toprule%
& & \multicolumn{5}{@{}c@{}}{CrossADR (Model 1) vs Ablated 1 (Model 2)} & \multicolumn{5}{@{}c@{}}{CrossADR (Model 1) vs Ablated 2 (Model 2)} \\
\cmidrule{3-7}\cmidrule{8-12}
Dataset & KG & Model 1 & Model 2 & P-Value & Sig & Effect Size & Model 1 & Model 2 & P-Value & Sig & Effect Size \\
\midrule
\multirow{4}{*}{CrossADR Dataset-D} 
& Basic      & 83.7\% & 82.4\% & $9.29\times10^{-6}$ & *** & 0.626 & 83.7\% & 81.8\% & $1.74\times10^{-6}$ & *** & 0.686 \\
& Ablation 1 & 68.8\% & 67.4\% & $1.22\times10^{-4}$ & *** & 0.450 & 68.8\% & 67.3\% & $8.56\times10^{-3}$ & ** & 0.351 \\
& Ablation 2 & 69.6\% & 67.6\% & $1.00\times10^{-3}$ & *** & 0.451 & 69.6\% & 65.4\% & $7.23\times10^{-9}$ & *** & 0.871 \\
& Ablation 3 & 65.5\% & 64.0\% & $1.31\times10^{-2}$ & * & 0.330 & 65.5\% & 62.1\% & $9.81\times10^{-5}$ & *** & 0.540 \\
\midrule
\multirow{4}{*}{CrossADR Dataset-R} 
& Basic      & 83.6\% & 81.7\% & $7.47\times10^{-12}$ & *** & 1.099 & 83.6\% & 80.9\% & $1.24\times10^{-10}$ & *** & 1.006 \\
& Ablation 1 & 68.4\% & 67.0\% & $1.36\times10^{-5}$  & *** & 0.465 & 68.4\% & 66.0\% & $2.64\times10^{-6}$  & *** & 0.613 \\
& Ablation 2 & 70.2\% & 69.1\% & $6.02\times10^{-5}$  & *** & 0.545 & 70.2\% & 66.0\% & $2.84\times10^{-9}$  & *** & 1.000 \\
& Ablation 3 & 65.1\% & 63.8\% & $1.00\times10^{-3}$  & *** & 0.368 & 65.1\% & 61.1\% & $5.55\times10^{-8}$  & *** & 0.961 \\
\botrule
\end{tabular*}
\end{table*}

To evaluate the contributions of the proposed modules within the CrossADR framework, a comprehensive ablation study is conducted. The analysis focuses on two primary architectural innovations: the cross-layer GNN on the Knowledge Graph (KG) and the cross-level attention module with a learnable ADR embedding space. Two ablated variants are developed for comparison. In Ablated 1, the cross-level attention module is replaced by the ADR association matrix approach used in OrganADR, while the cross-layer GNN module is retained. In Ablated 2, the cross-layer GNN is replaced with the traditional GNN structure found in OrganADR and EmerGNN, but the matrix-free learnable embedding module is maintained.

The quantitative results across different KGs and datasets are summarized in Table \ref{tab:sig_2}. The full CrossADR model (Model 1) consistently outperformed both ablated variants with high statistical significance. On the foundational "KG: Basic" of CrossADR Dataset-D, the ROC-AUC of CrossADR reached 83.7\%. In contrast, the performance of Ablated 1 and Ablated 2 dropped to 82.4\% ($P = 9.29 \times 10^{-6}$) and 81.8\% ($P = 1.74 \times 10^{-6}$), respectively. These results indicate that both the multi-scale knowledge flow and the associative learning mechanism are essential for optimal prediction.

A similar trend is observed in CrossADR Dataset-R. On the Basic KG, the ROC-AUC of the full model (83.6\%) is significantly higher than that of Ablated 1 (81.7\%) and Ablated 2 (80.9\%). Notably, the effect sizes for these comparisons reached 1.099 and 1.006, respectively. This suggests that the removal of either module leads to a substantial loss in the model's representative capability. 

The superiority of the full architecture is also maintained under information-sparse conditions. For instance, in the "KG: Ablation 2" scenario for Dataset-D, CrossADR achieved a mean ROC-AUC of 69.6\%. This represents a lead of 2.0\% over Ablated 1 (67.6\%) and 4.2\% over Ablated 2 (65.4\%). The performance degradation is most pronounced in Ablated 2 across most KGs, highlighting that the cross-layer GNN module plays a dominant role in capturing complex drug interactions. Overall, the ablation study confirms that the synergy between hierarchical feature integration and latent organ-level associations is fundamental to the robustness of CrossADR.

\begin{figure*}[!t]
\centering
\includegraphics[width=0.80\textwidth]{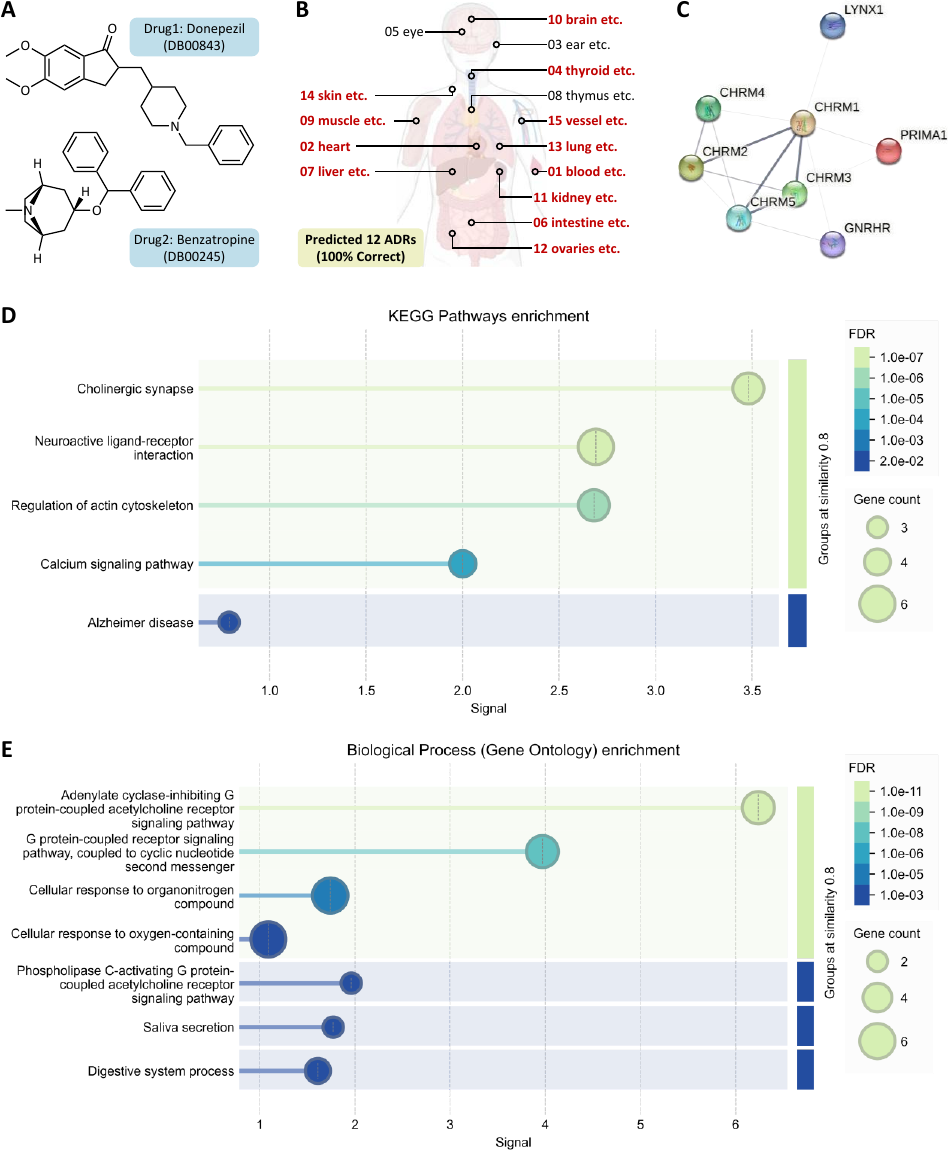}
\caption{Case study and interpretability analysis of CrossADR. (A) Drug combination to be analyzed: Drug 1, Donepezil and Drug 2, Benzatropine. (B) Predicted ADRs for the drug combination. (C) Visualization of the PPI network. The proteins in the PPI network are identified as important proteins by CrossADR. (D) KEGG pathway enrichment analysis of the key proteins. (E) Gene Ontology (GO) Biological Process enrichment analysis of the key proteins.}\label{fig8}
\end{figure*}

\subsection{Identification of key PPIs and pathways with CrossADR}

To demonstrate the practical utility and interpretability of the CrossADR framework, a case study is conducted on a representative drug combination: Donepezil (an acetylcholinesterase inhibitor) and Benzatropine (a muscarinic antagonist), as shown in Fig.\ref{fig8}A. This drug pair represents a classic scenario of cholinergic imbalance. In clinical practice, Donepezil is utilized to increase acetylcholine levels for the treatment of Alzheimer’s disease, while Benzatropine is administered to block acetylcholine receptors to alleviate extrapyramidal symptoms. The pharmacological conflict between these two agents—one acting as a ``cholinergic booster" and the other as a ``cholinergic blocker"— would create a severe "tug-of-war" within the autonomic and central nervous systems in human's body\cite{sink2008dual}.

As illustrated in Fig.\ref{fig8}B, the predictive capability of CrossADR is evaluated. For this specific drug pair, clinical records indicate the occurrence of adverse drug reactions (ADRs) in 12 distinct organs, including the brain, heart, liver, kidney, and intestine. The prediction results show that CrossADR accurately identifies the ADR status for all 12 organs, achieving $100\%$ accuracy. This high degree of alignment with clinical data suggests that the model effectively captures the ADRs arising from the antagonistic interaction of the two drugs.

To further investigate the mechanism underlying these multi-organ ADRs, CrossADR is utilized to prioritize the most influential proteins. Eight key proteins are identified: CHRM1, CHRM2, CHRM3, CHRM4, CHRM5, LYNX1, PRIMA1, and GNRHR. A PPI network is visualized in Fig.\ref{fig8}C, showing the dense functional connectivity between these targets. Notably, the muscarinic acetylcholine receptors (CHRM1–5) are identified as the core mediators, which is consistent with the known pharmacology where Benzatropine directly competes with the increased acetylcholine (facilitated by Donepezil) at these sites. Additionally, proteins like PRIMA1, which anchors acetylcholinesterase to neuronal membranes, and LYNX1, a modulator of nicotinic receptors, are identified as critical nodes in the protein-protein-interaction network\cite{perrier2002prima}.

The biological relevance of these identified proteins is further validated through enrichment analysis. As shown in the KEGG pathway enrichment (Fig.\ref{fig8}D), the "Cholinergic synapse" and "Neuroactive ligand-receptor interaction" pathways are the most significantly enriched. This matches the biological expectation that the ADRs are primarily driven by the disruption of cholinergic transmission. Furthermore, the Gene Ontology (GO) Biological Process analysis (Fig.\ref{fig8}E) reveals high enrichment in the "Adenylate cyclase-inhibiting G protein-coupled acetylcholine receptor signaling pathway" and "Phospholipase C-activating G protein-coupled acetylcholine receptor signaling pathway." These pathways directly correspond to the downstream signaling of M2/M4 and M1/M3/M5 receptors, respectively. The mechanisms identified by CrossADR are strongly supported by established pharmacological knowledge. For example, the disruption of M3 receptors in the intestine and bladder is known to cause gastrointestinal motility disorders and urinary retention\cite{kruse2012structure}, while the interference with M2 receptors in the heart is associated with conduction disturbances. The successful identification of these proteins and pathways demonstrates that CrossADR not only provides accurate organ-level predictions but also offers a high-resolution, biological interpretation of the molecular mechanisms driving complex ADRs with drug combination.

\section{Discussion}\label{sec4}

In this study, the CrossADR framework is introduced to address the challenge of organ-level adverse drug reaction prediction in combination pharmacotherapy. The model is rigorously evaluated on the newly constructed CrossADR-Dataset, which represents a significant expansion in drug combination space. Results indicate that CrossADR consistently achieves superior performance compared to state-of-the-art architectures and traditional machine learning methods across various knowledge graph configurations \cite{wan2025deepadr, li2026mhafr}. The effectiveness of the proposed architectural innovations is verified through comprehensive ablation studies and fine-grained cross-organ analysis. Furthermore, the practical utility of the framework is demonstrated through case studies that successfully identify key proteins and biological pathways, providing high-resolution interpretability for complex clinical outcomes \cite{gao2025precision, ren2025predicting}.

The primary innovation of CrossADR lies in the departure from traditional, fixed adverse drug reaction association matrices that often reflect inherent training data biases. Instead, a learnable ADR embedding space is implemented to capture organ-level information dynamically. This approach enhances cross-dataset generalization and flexibility by allowing the model to focus on interpretable biological knowledge rather than dataset-specific distributions \cite{zhang2026transformer, li2025llm}. Technically, the architecture emphasizes cross-layer feature integration within the gated-residual-flow module to fuse molecular features throughout the biomedical network. Cross-level associative learning is achieved through bi-directional cross-attention and gated mechanisms, which effectively bridge the gap between microscopic molecular signals and macroscopic organ-level responses.

The performance gains are attributed to the model's ability to mitigate deficiencies identified in previous architectures. Earlier models often treat molecular and organ information as isolated components or rely on simple attention that fails to capture the multi-scale evolution of knowledge flow. CrossADR addresses this by integrating vital information across different biological scales through deep feature fusion. By utilizing gated modules and residual connections, structural feature homogenization and numerical over-smoothing are prevented, ensuring that initial drug identities are preserved during deep propagation. Unlike models restricted by rigid, manually-curated correlation structures, the flexible associative learning mechanism allows for the identification of non-obvious clinical associations, such as those demonstrated in our cholinergic case study involving specific muscarinic receptor structures \cite{kruse2012structure, perrier2002prima, sink2008dual}.

Despite these advancements, several limitations are recognized that suggest directions for future work. The predictive accuracy remains partially dependent on the completeness and quality of the underlying biomedical knowledge graph \cite{tian2025ddinter, zheng2021pharmkg}. Additionally, the current framework is designed for binary classification and does not explicitly account for the severity or dose-dependent nature of adverse reactions. Future research will focus on the integration of single-cell data and longitudinal patient records to further refine personalized risk assessments \cite{langenberg2023biological, skou2022multimorbidity}. Enhancing the model to handle multi-modal inputs and temporal physiological changes is expected to improve the clinical utility of the system for real-world medication management and the training of next-generation bioinformatics experts \cite{jjingo2026pathways, chen2025evaluation}.

\section{Conclusion}

In this study, CrossADR is presented as a hierarchical framework designed for organ-level adverse drug reaction prediction through the integration of cross-layer feature fusion and cross-level associative learning. By utilizing a gated-residual-flow graph neural network and a learnable ADR embedding space, the model effectively captures the multi-scale dependencies between microscopic molecular interactions and macroscopic physiological responses across 15 organs. Evaluations on the comprehensive CrossADR-Dataset demonstrate that superior performance is consistently achieved over existing state-of-the-art methods in diverse clinical scenarios. Furthermore, the biological interpretability of the framework is confirmed through the identification of key protein-protein interactions and metabolic pathways, suggesting that CrossADR serves as a robust and reliable tool for enhancing safety assessments in combination pharmacotherapy and supporting precision clinical decision-making, drug safety as well as drug management.

\section{Conflicts of interest}
The authors declare that they have no competing interests.






\bibliographystyle{unsrt}
\bibliography{reference}

\end{document}